\documentclass{article}

\pdfoutput=1
\usepackage{arxiv}

\usepackage[utf8]{inputenc} 
\usepackage[T1]{fontenc}    
\usepackage{hyperref}       
\usepackage{url}            
\usepackage{booktabs}       
\usepackage{amsfonts}       
\usepackage{nicefrac}       
\usepackage{microtype}      
\usepackage{lipsum}
\usepackage{graphicx}
\usepackage{multirow}
\usepackage{bm} 

\graphicspath{ {./images/} }

\title{Resolution Invariant Autoencoder}

\author{
 Ashay Patel \\
  School of Biomedical Engineering \\
  and Imaging Sciences \\
  King's College London\\
  London \\
  \texttt{ashay.patel@kcl.ac.uk} \\
   \And
 Michela Antonelli \\
  School of Biomedical Engineering \\
    and Imaging Sciences \\
  King's College London\\
  London \\
  \texttt{michela.antonelli@kcl.ac.uk} \\
  \And
 Sebastien Ourselin \\
  School of Biomedical Engineering \\
    and Imaging Sciences \\
  King's College London\\
  London \\
  \texttt{sebastien.ourselin@kcl.ac.uk} \\
  \And
   M. Jorge Cardoso \\
  School of Biomedical Engineering \\
    and Imaging Sciences \\
  King's College London\\
  London \\
  \texttt{m.jorge.cardoso@kcl.ac.uk} \\
}

\begin{document}
\maketitle
\begin{abstract}
Deep learning has significantly advanced medical imaging analysis, yet variations in image resolution remain an overlooked challenge. Most methods address this by resampling images, leading to either information loss or computational inefficiencies. While solutions exist for specific tasks, no unified approach has been proposed. We introduce a resolution-invariant autoencoder that adapts spatial resizing at each layer in the network via a learned variable resizing process, replacing fixed spatial down/upsampling at the traditional factor of 2. This ensures a consistent latent space resolution, regardless of input or output resolution. Our model enables various downstream tasks to be performed on an image latent whilst maintaining performance across different resolutions, overcoming the shortfalls of traditional methods. We demonstrate its effectiveness in uncertainty-aware super-resolution, classification, and generative modelling tasks and show how our method outperforms conventional baselines with minimal performance loss across resolutions.
\end{abstract}

\section{Introduction}

Deep learning has seen significant advancements in natural and medical imaging, excelling in tasks like classification, segmentation, and generative modelling. However, most medical imaging research focuses on performance in specific use cases, assuming uniform data acquisition enforced by resampling to standardized voxel sizes \cite{Valiant1984,Ben-David2006,Varoquaux2022}. In the field of medical imaging a defined voxel spacing ($mm^3$) is used to define the image resolution, an essential piece of information for relative pre-processing applications whilst ensuring consistency for convolutional models. Yet, downsampling high-resolution images causes information loss, while upsampling low-resolution ones can waste resources. This common pre-processing, seen in challenges like BRATS \cite{Menze2014}, Medical Segmentation Decathlon \cite{Antonelli2021}, and the WMH Segmentation Challenge \cite{Kuijf2019}, can hinder generalization and increase processing time \cite{Decuyper2021,Dinsdale2022,Varoquaux2022}. Our model addresses this by ensuring resolution invariance.

To our knowledge, no comprehensive solution exists for handling image resolution variances across all image processing tasks. Instead, most research focuses on problem-specific approaches with effective but limited further applications. One common method is training models on images of varying resolutions without architectural modifications. SynthSeg ~\cite{Billot2023} adopts this approach by using a generative model to create synthetic images at multiple resolutions, followed by CNN-based segmentation. While this improves robustness, performance degrades at lower resolutions, and often models trained on a single resolution outperform the multi-resolution approach, suggesting a trade-off between generalization and state-of-the-art performance.

An alternative approach to resolution invariance is found in ~\cite{Patel2023}, which employs a two-part model combining a Vector Quantized Variational Autoencoder and a Transformer with a spatial conditioning mechanism. This mechanism, using a three-channel coordinate system, encodes resolution, field-of-view, and orientation, allowing the model to adapt to arbitrary resolutions. Similarly, ~\cite{Patel2024} integrates coordinate channels into a Latent Diffusion Model to condition image generation on resolution and field-of-view. While effective, these methods require architectural modifications and training adaptations, limiting their general applicability across all imaging tasks without significant rework. It is also unseen whether simply concatenating channels is sufficient conditioning for convolutional networks for segmentation and classification that do not contain any attention mechanisms.

The work in ~\cite{Joutard2024} proposed using hypernetworks for image segmentation, where subnetworks of linear layers generate the weights for a U-Net’s convolutions. By taking resolution as input, these subnetworks adjust the U-Net’s weights to accommodate resolution changes. This method outperforms standard U-Nets when training and testing data vary in resolution. However, hypernetworks can be computationally demanding, especially for larger models, as each convolution requires a dedicated subnetwork. This makes the approach impractical for tasks like generative modelling, where larger models are used.

While prior works offer insights into handling resolution variance, no single solution provides a simple, effective, and flexible approach across all imaging tasks. To address this, we propose a resolution-invariant autoencoder. Its key innovation is variable downsampling and upsampling, replacing the fixed spatial resizing by a factor of 2 used in most networks. This allows encoding images into a latent space of fixed resolution, enabling downstream tasks to operate on the latent without architectural or training modifications. Additionally, by incorporating KL regularization and estimating information loss between resolutions, we can apply noise to the latent in proportion to the estimated information loss so that our model quantifies uncertainty when encoding low-resolution images to higher resolutions.

From training such a model, we show how it can be used for a multitude of applications including image super-resolution with associated uncertainty quantification, image classification and generative modelling. This is however a non-exhaustive list and the applications of such a model are far reaching to any model that can perform analysis on an image latent. During testing we show how data of varying resolutions can be used seamlessly with this model for any of these tasks without any image pre-processing or performance degradation.

\section{Methodology}

In medical imaging, where high-dimensional data poses computational challenges, working in a latent space offers a more efficient alternative. Thus, autoencoders provide an optimal framework for encoding data into a uniform resolution format, facilitating diverse downstream imaging tasks. Autoencoders (AEs) consist of an encoder and decoder, where the encoder maps an input image to a lower-dimensional manifold $z$, and the decoder reconstructs the image: $x' = Dec(z)$. Their typical structure involves progressive downsampling at a spatial factor of 2 along each dimension in the encoder and upsampling in the decoder. A key limitation of AEs is the lack of latent space regularization. Variational autoencoders (VAEs) ~\cite{Kingma2013} address this by enforcing a probabilistic structure on the latent space. Autoencoders are widely used with downstream models for tasks such as generative modelling \cite{Rombach2022,Pinaya2022,Tudosiu2024}, segmentation ~\cite{Quoc2023}, super-resolution ~\cite{Luo2023}, and anomaly detection/out-of-distribution detection \cite{Patel2022,Graham2023,Pinaya2022_AD}. 

\subsection{Learnt Variable Resampling}

\begin{figure}
\centering
\includegraphics[width=14cm]{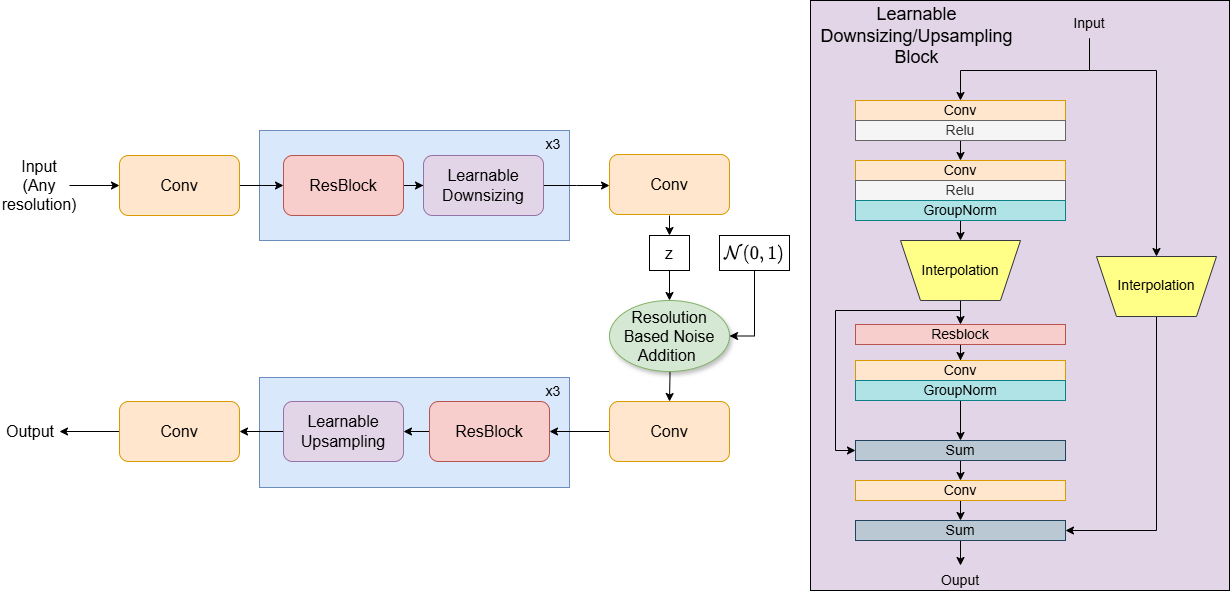}
\caption{Illustration of (left) AE architecure consisting of 3 layers of residual blocks followed by variable resizing blocks along the encoder and decoder, (right) architecture for learnable variable resizing block} \label{learnable_resampling}
\end{figure}

The key component enabling resolution invariance in our autoencoder is a learnable downsizing block, inspired by ~\cite{Talebi2021}. This approach was originally used as a preprocessing step to downsample high-resolution images into fixed low-resolution representations with additional channels for faster more efficient classification. Unlike traditional methods relying on strided convolutions, pooling, transposed convolutions or fixed-factor interpolation, this block performs arbitrary interpolation to any target resolution. Additionally, a skip-connection mechanism integrates the resized input with new convolutional features. We configure this however so it can be used for both upsampling and downsampling. An illustration of this and the overall model architecture is shown in Figure \ref{learnable_resampling}.

We integrate this learnable resampling block into an autoencoder to produce a latent space of fixed resolution. Instead of traditional downsampling and upsampling by a factor of 2, we adjust the resampling scale based on input resolution. For instance, if the highest expected resolution is $1mm^3$ isotropic, a standard autoencoder with three downsampling layers would yield an effective latent resolution of $(2^3)mm^3 = 8mm^3$ isotropic. If the input resolution is $2mm^3$ isotropic, we adjust the resizing factor to $\sim1.58$ per layer, ensuring the latent remains $8mm^3$ isotropic.

For a three-layer autoencoder, defining the latent resolution as the highest training resolution downsampled by a factor of 2 per layer (as in a standard autoencoder) allows us to compute the necessary downsampling or upsampling ratio for any given input as follows:

\begin{equation}\label{eq:resampling_factor}
d_i = (l_i/r_i)^{1/n}
\end{equation}

where $d$ is the downsampling/upsampling factor, $l$ is the latent resolution, $r$ is the resolution of the original image (encoding) or output image (decoding), $i$ refers to the x,y or z dimension and $n$ is the number of layers in the autoencoder. Thus, the autoencoder can encode data at any resolution and reconstruct outputs at any resolution, as the downsampling and upsampling factors are independent and treated as non-parametric inputs.

We use the latent of the highest resolution training samples to define the fixed latent resolution so there is no information loss from over downsampling the data. 

\subsection{Latent Consistency Loss}
To enhance resolution invariance, we train the model to ensure that latents from the same image at different resolutions remain similar. This is achieved using a latent consistency loss ($L_{latent}$). During training, a high-resolution image is downsampled to a random lower resolution, and both versions are encoded (to the same fixed size). An $L_1$ loss is applied between the latents to enforce consistency. The high-resolution latent is decoded to a high-resolution image, whilst the low-resolution latent is decoded to both its original resolution and the high-resolution version. Reconstruction losses, combining $L_1$ and perceptual loss (via LPIPS ~\cite{Zhang2018}), are applied. To enhance realism, an adversarial loss is applied across all reconstructions. Additionally, a Kullback–Leibler divergence loss $L_{KL}$, regularizes the latent space to prevent unbounded variances.

An overview of the training framework is shown in Figure \ref{training_schema}.

\begin{figure}
\centering
\includegraphics[width=12cm]{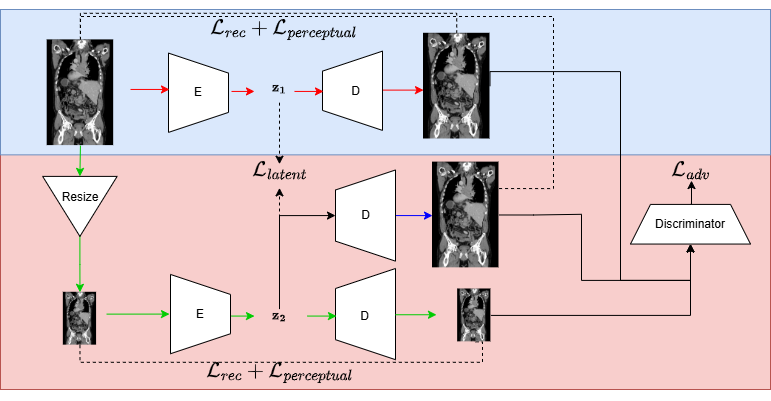}
\caption{Training pipeline indicating flow of data during training of the resolution invariant AE} \label{training_schema}
\end{figure}
 
\subsection{Information Loss as Uncertainty}

The second model adaptation focuses on structuring the latent space for encoding data at different resolutions. While true resolution invariance is impossible due to lost details at lower resolutions, information can remain invariant at different frequencies. Low-frequency details persist, while high-frequency ones may be lost. To account for this, we inject noise into the latent space based on expected information loss, estimated experimentally using the Structural Similarity Index Measure (SSIM) ~\cite{Wang2004}. By comparing high-resolution images downsampled to various lower resolutions (up to a factor of 6 along each dimension) and then upsampled, we calculate SSIM drop as an estimate of information loss, which is averaged over a training subset (20 samples) and used to adjust the latent encoding. This yields a resolution-dependent noise estimate applicable to all samples. Using this, when encoding a lower-resolution image, the latent is reconfigured as follows:

\begin{equation}\label{eq:noised_latent}
z_{noised} = (1-\gamma)z + \gamma \mathcal{N}(0,1)
\end{equation}

Where $\gamma$ represents the estimated information loss and $z$ corresponds to the original encoded latent. By introducing noise proportional to $\gamma$, we account for the uncertainty inherent in reconstructing high-resolution images from lower-resolution inputs. This approach enables both image super-resolution and the quantification of super-resolution uncertainty, as the lost information is replaced with noise within the latent space. Full implementation details can be seen with the submitted open source code repository.

\section{Experiments}
To demonstrate the utility of the resolution-invariant autoencoder, we conducted three experiments across different medical imaging scenarios. These experiments highlight the autoencoder's adaptability to various tasks without compromising model performance, even when data originates from different resolutions. Specifically, we evaluate its use in: super-resolution at any arbitrary resolutions, classification with training and testing data at different resolutions, and generative modelling with multi-resolution training data. This is not an exhaustive list, as the autoencoder's applicability extends to any imaging analysis task.

\subsection{Experiment 1: Whole Body CT Super-resolution}
The first experiment focuses on image super-resolution. Since our autoencoder encodes data from any resolution to a fixed-resolution latent space and can decode to any target resolution, it can naturally enable a super-resolution task at any resolution factor. We also qualitatively assess the impact of resolution-based noise injection to quantify uncertainty in the super-resolved images. Few works, such as \cite{Hu2019}, allow super-resolution from arbitrary input to output resolutions; this method that serves as our baseline used a U-Net architecture and upsamples low-resolution images to match high-resolution ground truth dimensions before being fed through the network. We test both models across various lower resolutions, super-resolving to the highest resolution in the training data. For this, we use whole-body CT data from \cite{Gatidis2022}, comprising 897 high-resolution whole-body CT scans, resampled to $1.6 \times 1.6 \times 2.5 mm^3$ for computational efficiency.

Results are summarized in Table \ref{tab:super_resolution_results}, with qualitative comparisons shown in Figure 3. Performance is measured using Peak Signal-to-Noise Ratio (PSNR) and the Fréchet Inception Distance (FID) as in ~\cite{Heusel2017} using Med3d as a feature extractor ~\cite{Chen2019}. To assess the uncertainty of reconstructions each image is super-resolved 40 times, and we compute the standard deviation across the reconstructions to generate an uncertainty map.

{\Large
\begin{table}[]
\centering
\caption{Results for super-resolution at varying resolutions for invariant AE and U-Net based approach. Statistically significant findings ($p < .01$) reported in bold.}
\label{tab:super_resolution_results}
\resizebox{\textwidth}{!}{%
\begin{tabular}{|cl|cccccccc|}
\hline
                                            & \multicolumn{1}{c|}{}   & \multicolumn{8}{c|}{Resolution Change Factor}                                                                          \\ \cline{3-10} 
                                            & \multicolumn{1}{c|}{}   & \multicolumn{1}{c|}{1.2$\times$}                                                                & \multicolumn{1}{c|}{1.6$\times$}                                                                & \multicolumn{1}{c|}{2.0$\times$}                                                       & \multicolumn{1}{c|}{2.4$\times$}                                                                & \multicolumn{1}{c|}{2.8$\times$}                                                                & \multicolumn{1}{c|}{3.2$\times$}                                                                & \multicolumn{1}{c|}{3.6$\times$}                                                                & \multicolumn{1}{c|}{4.0$\times$}                                                                \\ \hline
\multicolumn{1}{|c|}{\multirow{2}{*}{FID $\downarrow$}}  & U-Net                   & \multicolumn{1}{c|}{$\bm{0.0642 \pm 0.0027}$} & \multicolumn{1}{c|}{$0.0729 \pm 0.0029$} & \multicolumn{1}{c|}{$0.0772 \pm 0.0022$} & \multicolumn{1}{c|}{$0.0832 \pm 0.0025$}          & \multicolumn{1}{c|}{$0.0881 \pm 0.0024$}          & \multicolumn{1}{c|}{$0.0860 \pm 0.0029$}          & \multicolumn{1}{c|}{$0.0917 \pm 0.0031$}          & $0.1161 \pm 0.0035$          \\ \cline{2-10} 
\multicolumn{1}{|c|}{}                      & Resolution Invariant AE & \multicolumn{1}{c|}{$0.0715 \pm 0.0032$}          & \multicolumn{1}{c|}{$0.0746 \pm 0.0036$}          & \multicolumn{1}{c|}{$0.0760 \pm 0.0034$} & \multicolumn{1}{c|}{$\bm{0.0774 \pm 0.0038}$} & \multicolumn{1}{c|}{$\bm{0.0810 \pm 0.0041}$} & \multicolumn{1}{c|}{$\bm{0.0803 \pm 0.0039}$} & \multicolumn{1}{c|}{$\bm{0.0840 \pm 0.0037}$} & $\bm{0.0812 \pm 0.0043}$ \\ \hline
\multicolumn{1}{|c|}{\multirow{2}{*}{PSNR $\uparrow$}} & U-Net                   & \multicolumn{1}{c|}{$28.41 \pm 2.33$}             & \multicolumn{1}{c|}{$27.62 \pm 2.94$}              & \multicolumn{1}{c|}{$26.24 \pm 2.76$}    & \multicolumn{1}{c|}{$26.88 \pm 3.19$}             & \multicolumn{1}{c|}{$24.28 \pm 2.74$}             & \multicolumn{1}{c|}{$23.56 \pm 3.12$}             & \multicolumn{1}{c|}{$21.15 \pm 2.74$}             & $19.90 \pm 2.67$             \\ \cline{2-10} 
\multicolumn{1}{|c|}{}                      & Resolution Invariant AE & \multicolumn{1}{c|}{$27.85 \pm 3.43$}             & \multicolumn{1}{c|}{$27.05 \pm 3.38$}             & \multicolumn{1}{c|}{$26.36 \pm 3.35$}    & \multicolumn{1}{c|}{$26.40 \pm 3.24$}             & \multicolumn{1}{c|}{$\bm{26.14 \pm 3.11}$}    & \multicolumn{1}{c|}{$\bm{26.01 \pm 3.17}$}    & \multicolumn{1}{c|}{$\bm{25.76 \pm 3.19}$}    & $\bm{25.64 \pm 3.04}$    \\ \hline
\end{tabular}%
}
\end{table}}

\begin{figure}
\centering
\includegraphics[width=14cm]{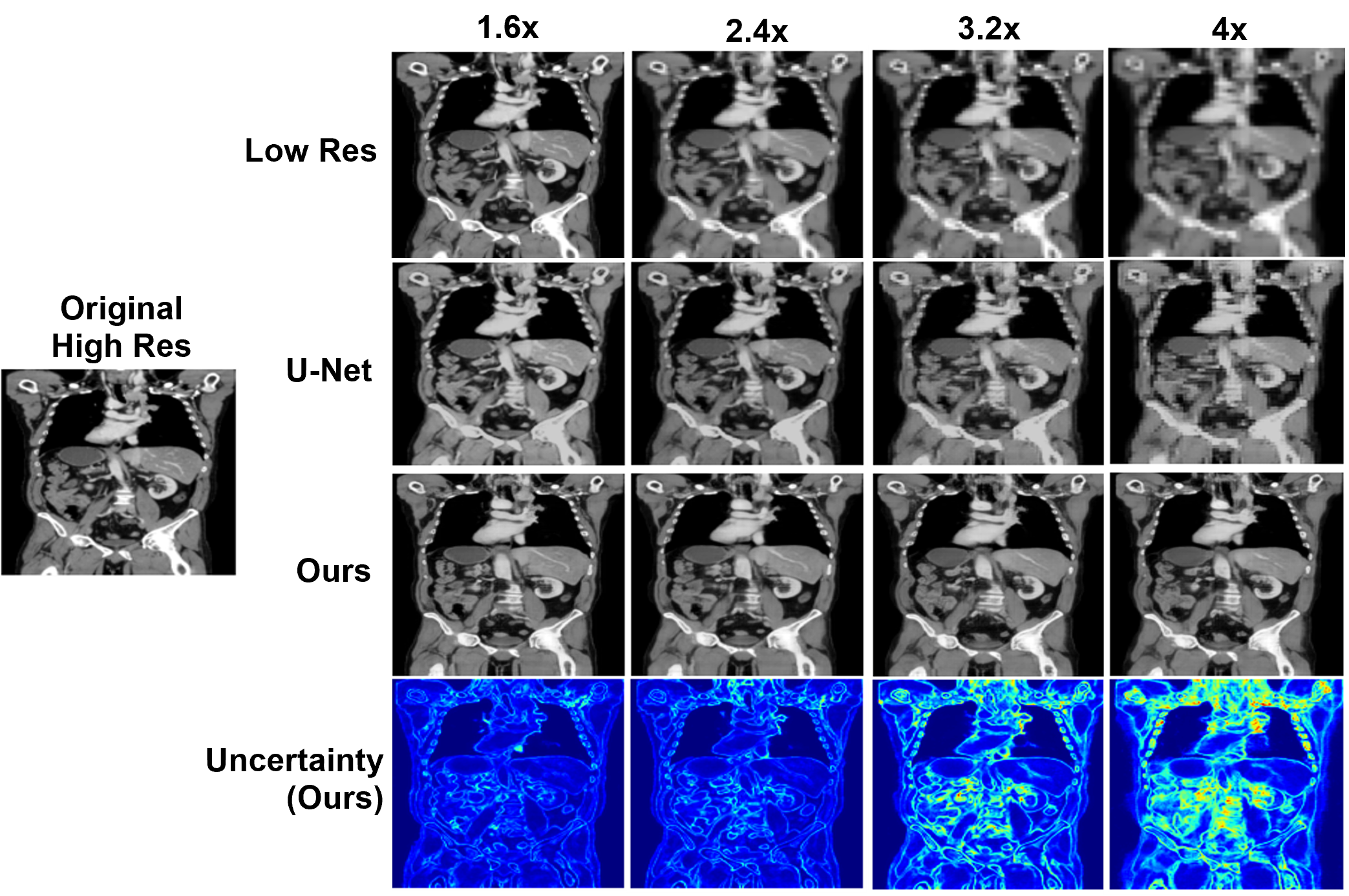}
\caption{Qualitative results showcasing (left) the original ground truth image, (top) low resolution input image ( resampled to same output size via linear interpolation), (2nd row) super-resolved image using the U-Net approach, (3rd row) super-resolved image using our proposed method, (4th row) the uncertainty for our approach.} \label{training_schema}
\end{figure}

The results from the resolution-invariant model align well with expectations regarding uncertainty across different resolutions. At higher resolutions, with minimal information loss, the model reconstructs the high-resolution image with little uncertainty and better quantitative metrics. However, as input resolution decreases, uncertainty increases due to greater information loss. This effect is particularly noticeable in regions with finer structural details, where lower-resolution inputs struggle to preserve high-frequency information. Unlike the U-Net approach with visible performance drops at high super-resolution ratios, our approach visibly shows a stable performance across all resolution factors with little performance degradation even at the most extreme factors of 4 across each dimension (64 volume wise). This highlights the benefit of the consistent latent resolution showing statistically superior performance across most super-resolution factors.

\subsection{Experiment 2: ADNI Alzheimer's Classification}

\begin{table}[]
\centering
\caption{AUROC for classifiers trained on image latents using a Autoencoder KL with image resampling or our resolution invariant AE}
\label{tab:classification_results}
\resizebox{\textwidth}{!}{%
\begin{tabular}{|c|ccc|ccc|}
\hline
           & \multicolumn{3}{c|}{Autoencoder KL}                                                     & \multicolumn{3}{c|}{AE-KL Invariant}                                                 \\ \cline{2-7} 
           & \multicolumn{1}{c|}{HR Training} & \multicolumn{1}{c|}{LR Training} & Mixed Training & \multicolumn{1}{c|}{HR Training} & \multicolumn{1}{c|}{LR Training} & Mixed Training \\ \hline
HR Testing & \multicolumn{1}{c|}{0.78}        & \multicolumn{1}{c|}{0.69}        & 0.67           & \multicolumn{1}{c|}{0.78}        & \multicolumn{1}{c|}{0.74}        & 0.77           \\ \hline
LR Testing & \multicolumn{1}{c|}{0.68}        & \multicolumn{1}{c|}{0.72}        & 0.66           & \multicolumn{1}{c|}{0.76}        & \multicolumn{1}{c|}{0.77}        & 0.74           \\ \hline
\end{tabular}%
}
\end{table}

In the second experiment, we assess the resolution-invariant AE for classification using the ADNI dataset ~\cite{ADNI}, which includes 1134 MRI brain scans at $1\times 1 \times 1.2 mm^3$ resolution categorized as Cognitively Normal, Mild Cognitive Impairment, or Alzheimer’s Disease. A classifier is trained on encoded latents from two autoencoders: our resolution-invariant AE and a baseline Autoencoder KL ~\cite{Rombach2022}, which lacks variable resizing. Both AEs are frozen post-training, serving only for encoding. Models are trained on high-resolution, low-resolution, or mixed-resolution data. Since the ADNI dataset is high-resolution, lower resolutions are simulated via resampling. The resolution-invariant AE maintains a consistent latent space without pre-processing, while the Autoencoder KL requires upsampling for lower resolution inputs. Performance is evaluated on high and low-resolution test sets using AUROC, with results shown in Table \ref{tab:classification_results}.

The results indicate that the classifier trained with the resolution-invariant AE performs as well as or better than the Autoencoder KL with image resampling. The performance gap is most notable when training and testing involve mixed resolutions or mismatched resolutions (e.g., low-resolution training and high-resolution testing). This highlights the AE’s adaptability to varying resolutions without requiring pre-processing. Even in the low-resolution-only scenario, our approach outperforms the Autoencoder KL, likely due to a more information-rich latent space aligned with high-resolution data through the training schema.

\subsection{Experiment 3: Whole Body CT Generative Modelling}

The final experiment demonstrates how the resolution-invariant AE can address data scarcity in generative modelling by utilizing data of any resolution. We train a Latent Diffusion Model (LDM) \cite{Rombach2022} on whole-body CT data (from Experiment 1). The resolution-invariant AE encodes the training data where we then experiment three training scenarios: 1) all high-resolution data, 2) limited high-resolution samples, and 3) a mix of few high-resolution and many low-resolution samples. The goal is to show that by using the AE we can supplement high-resolution data ($1.6 \times 1.6 \times 2.5mm^3$) with low-resolution data ($3.2 \times 3.2 \times 5mm^3$), narrowing the performance gap to an ideal scenario. We generate 100 high-res samples for each model and evaluate their performance using the FID metric.

\begin{table}[]
\centering
\caption{FID scores for different training scenarios: few high-resolution samples, few high-resolution samples supplemented with many low-resolution samples, and an oracle dataset (many high-resolution samples).}
\label{tab:LDM_results}
\resizebox{10cm}{!}{%
\begin{tabular}{|c|c|c|c|}
\hline
Training Set  & 50 HR Samples & 50 HR + 797 LR Samples & 847 HR Samples \\ \hline
FID $\downarrow$ & $0.2238 \pm 0.019$ & $0.1606 \pm 0.011$ & $0.1408 \pm 0.008$ \\ \hline
\end{tabular}%
}
\end{table}

As expected, the model trained with many high-resolution samples performs best. However, the key advantage of our autoencoder appears when trained with few high-resolution and many low-resolution samples. While performance decreases with only 50 high-resolution samples, supplementing many low-resolution samples substantially narrows the performance gap compared to a model trained solely on high-resolution data. This demonstrates the utility of our model in leveraging all data, even when a gold standard training set is unavailable.

\section{Conclusion}
This work demonstrates how reconfiguring an autoencoder’s architecture and training pipeline enables the generation of a resolution-invariant latent representation. We believe this approach has broad applications as shown in our experiments. By providing a universal method for handling data of varying resolutions, our model enhances flexibility and removes the need to discard low-resolution data or rely on time-consuming resampling. We hope this study encourages further research into this crucial yet underexplored area.

\newpage
%
%
%
%

\end{document}